\documentclass[10pt,twocolumn,letterpaper]{article}

\usepackage{cvpr}
\usepackage{times}
\usepackage{epsfig}
\usepackage{graphicx}
\usepackage{amsmath}
\usepackage{amssymb}
\usepackage[linesnumbered,ruled]{algorithm2e}
\usepackage{verbatim}
\usepackage{mathtools}
\usepackage{tabularx}
\usepackage{multirow}
\newtheorem{theorem}{Theorem}
\newtheorem{lemma}[theorem]{Lemma}

\usepackage[pagebackref=true,breaklinks=true,letterpaper=true,colorlinks,bookmarks=false]{hyperref}

\cvprfinalcopy 

\def\cvprPaperID{4220} 

\ifcvprfinal\fi
\begin{document}

\title{PointConv: Deep Convolutional Networks on 3D Point Clouds}

\author{Wenxuan Wu, Zhongang Qi, Li Fuxin\\
CORIS Institute, Oregon State University\\
{\tt\small wuwen, qiz, lif@oregonstate.edu}
}

\maketitle

\begin{abstract}
Unlike images which are represented in regular dense grids, 3D point clouds are irregular and unordered, hence applying convolution on them can be difficult. In this paper, we extend the dynamic filter to a new convolution operation, named PointConv. PointConv can be applied on point clouds to build deep convolutional networks. We treat convolution kernels as nonlinear functions of the local coordinates of 3D points comprised of weight and density functions. With respect to a given point, the weight functions are learned with multi-layer perceptron networks and density functions through kernel density estimation. The most important contribution of this work is a novel reformulation proposed for efficiently computing the weight functions, which allowed us to dramatically scale up the network and significantly improve its performance. The learned convolution kernel can be used to compute translation-invariant and permutation-invariant convolution on any point set in the 3D space. 
Besides, PointConv can also be used as deconvolution operators to propagate features from a subsampled point cloud back to its original resolution. Experiments on ModelNet40, ShapeNet, and ScanNet show that deep convolutional neural networks built on PointConv are able to achieve state-of-the-art on challenging semantic segmentation benchmarks on 3D point clouds. Besides, our experiments converting CIFAR-10 into a point cloud showed that networks built on PointConv can match the performance of convolutional networks in 2D images of a similar structure.
\end{abstract}

\vspace{-0.2in}
\section{Introduction}

In recent robotics, autonomous driving and virtual/augmented reality applications, sensors that can directly obtain 3D data are increasingly ubiquitous. This includes indoor sensors such as laser scanners, time-of-flight sensors such as the Kinect, RealSense or Google Tango, structural light sensors such as those on the iPhoneX, as well as outdoor sensors such as LIDAR and MEMS sensors. The capability to directly measure 3D data is invaluable in those applications as depth information could remove a lot of the segmentation ambiguities from 2D imagery, and surface normals provide important cues of the scene geometry. 

In 2D images, convolutional neural networks (CNNs) have fundamentally changed the landscape of computer vision by greatly improving results on almost every vision task. CNNs succeed by utilizing translation invariance, so that the same set of convolutional filters can be applied on all the locations in an image, reducing the number of parameters and improving generalization. We would hope such successes to be transferred to the analysis of 3D data. However, 3D data often come in the form of point clouds, which is a set of unordered 3D points, with or without additional features (e.g. RGB) on each point. Point clouds are unordered and do not conform to the regular lattice grids as in 2D images. It is difficult to apply conventional CNNs on such unordered input. An alternative approach is to treat the 3D space as a volumetric grid, but in this case, the volume will be sparse and CNNs will be computationally intractable on high-resolution volumes.


In this paper, we propose a novel approach to perform convolution on 3D point clouds with non-uniform sampling. We note that the convolution operation can be viewed as a discrete approximation of a continuous convolution operator. In 3D space, we can treat the weights of this convolution operator to be a (Lipschitz) continuous function of the local 3D point coordinates with respect to a reference 3D point. The continuous function can be approximated by a multi-layer perceptron(MLP), as done in  \cite{simonovsky2017dynamic} and \cite{jia2016dynamic}. But these algorithms did not take non-uniform sampling into account. 
We propose to use an inverse density scale to re-weight the continuous function learned by MLP, which corresponds to the Monte Carlo approximation of the continuous convolution. We call such an operation \textbf{PointConv}. PointConv involves taking the positions of point clouds as input and learning an MLP to approximate a weight function, as well as applying a inverse density scale on the learned weights to compensate the non-uniform sampling.

The naive implementation of PointConv is memory inefficient when the channel size of the output features is very large and hence hard to train and scale up to large networks. In order to reduce the memory consumption of PointConv, we introduce an approach which is able to greatly increase the memory efficiency using a reformulation that changes the summation order. The new structure is capable of building multi-layer deep convolutional networks on 3D point clouds that have similar capabilities as 2D CNN on raster images. We can achieve the same translation-invariance as in 2D convolutional networks, and the invariance to permutations on the ordering of points in a point cloud. 

In segmentation tasks, the ability to transfer information gradually from coarse layers to finer layer is important. Hence, a deconvolution operation \cite{noh2015learning} that can fully leverage the feature from a coarse layer to a finer layer is vital for the performance. Most state-of-the-art algorithms \cite{qi2017pointnet,qi2017pointnet++} are unable to perform deconvolution, which restricts their performance on segmentation tasks. Since our PointConv is a full approximation of convolution, it is natural to extend PointConv to a PointDeconv, which can fully untilize the information in coarse layers and propagate to finer layers. By using PointConv and PointDeconv, we can achieve improved performance on semantic segmentation tasks.

The contributions of our work are:

$ \bullet $ We propose PointConv, a density re-weighted convolution, which is able to fully approximate the 3D continuous convolution on any set of 3D points. 

$ \bullet $ We design a memory efficient approach to implement PointConv using a change of summation order technique, most importantly, allowing it to scale up to modern CNN levels. 

$ \bullet $ We extend our PointConv to a deconvolution version(PointDeconv) to achieve better segmentation results.

Experiments show that our deep network built on PointConv is highly competitive against other point cloud deep networks and achieve state-of-the-art results in part segmentation \cite{chang2015shapenet} and indoor semantic segmentation benchmarks \cite{dai2017scannet}. In order to demonstrate that our PointConv is indeed a true convolution operation, we also evaluate PointConv on CIFAR-10 by converting all pixels in a 2D image into a point cloud with 2D coordinates along with RGB features on each point. Experiments on CIFAR-10  show that the classification accuracy of our PointConv is comparable with a image CNN of a similar structure, far outperforming previous best results achieved by point cloud networks. As a basic approach to CNN on 3D data, we believe there could be many potential applications of PointConv.

\section{Related Work}

Most work on 3D CNN networks convert 3D point clouds to 2D images or 3D volumetric grids. \cite{su2015multi,qi2016volumetric} proposed to project 3D point clouds or shapes into several 2D images, and then apply 2D convolutional networks for classification. Although these approaches have achieved dominating performances on shape classification and retrieval tasks, it is nontrivial to extend them to high-resolution scene segmentation tasks \cite{dai2017scannet}. \cite{wu20153d,maturana2015voxnet,qi2016volumetric} represent another type of approach that voxelizes point clouds into volumetric grids by quantization and then apply 3D convolution networks. This type of approach is constrained by its 3D volumetric resolution and the computational cost of 3D convolutions. \cite{riegler2017octnet} improves the resolution significantly by using a set of unbalanced octrees where each leaf node stores a pooled feature representation. Kd-networks\cite{klokov2017escape} computes the representations in a feed-forward bottom-up fashion on a Kd-tree with certain size. In a Kd-network, the input number of points in the point cloud needs to be the same during training and testing, which does not hold for many tasks. SSCN \cite{graham2017submanifold} utilizes the convolution based on a volumetric grid with novel speed/memory improvements by considering CNN outputs only on input points. However, if the point cloud is sampled sparsely, especially when the sampling rate is uneven, for the sparsely sampled regions on may not be able to find any neighbor within the volumetric convolutional filter, which could cause significant issues. 

Some latest work \cite{ravanbakhsh2016deep,qi2017pointnet,qi2017pointnet++, su2018splatnet, tatarchenko2018tangent, hua2018pointwise, groh2018flex, verma2018feastnet} directly take raw point clouds as input without converting them to other formats. \cite{qi2017pointnet,ravanbakhsh2016deep} proposed to use shared multi-layer perceptrons and max pooling layers to obtain features of point clouds. Because the max pooling layers are applied across all the points in point cloud, it is difficult to capture local features. PointNet++ \cite{qi2017pointnet++} improved the network in PointNet \cite{qi2017pointnet} by adding a hierarchical structure. The hierarchical structure is similar to the one used in image CNNs, which extracts features starting from small local regions and gradually extending to larger regions. The key structure used in both PointNet \cite{qi2017pointnet} and PointNet++ \cite{qi2017pointnet++} to aggregate features from different points is max-pooling. However, max-pooling layers keep only the strongest activation on features across a local or global region, which may lose some useful detailed information for segmentation tasks. \cite{su2018splatnet} presents a method that projects the input features of the point clouds onto a high-dimensional lattice, and then apply bilateral convolution on the high-dimensional lattice to aggregate features, which called ``SPLATNet". The SPLATNet \cite{su2018splatnet} is able to give comparable results as PointNet++ \cite{qi2017pointnet++}. The tangent convolution \cite{tatarchenko2018tangent} projects local surface geometry on a tangent plane around every point, which gives a set of planar-convolutionable tangent images. The pointwise convolution \cite{hua2018pointwise} queries nearest neighbors on the fly and bins the points into kernel cells, then applies kernel weights on the binned cells to convolve on point clouds. Flex-convolution \cite{groh2018flex} introduced a generalization of the conventional convolution layer along with an efficient GPU implementation, which can applied to point clouds with millions of points. FeaStNet \cite{verma2018feastnet} proposes to generalize conventional convolution layer to 3D point clouds by adding a soft-assignment matrix. PointCNN \cite{li2018pointcnn} is to learn a $ \chi -$transformation from the input points and then use it to simultaneously weight and permute the input features associated with the points. Comparing to our approach, PointCNN is unable to achieve permutation-invariance, which is desired for point clouds.

The work \cite{simonovsky2017dynamic,jia2016dynamic, wang2018dynamic, hermosilla2018monte, wang2018deep} and \cite{xu2018spidercnn} propose to learn continuous filters to perform convolution. \cite{jia2016dynamic} proposed that the weight filter in 2d convolution can be treated as a continuous function, which can be approximated by MLPs. \cite{simonovsky2017dynamic} firstly introduced the idea into 3d graph structure. \cite{wang2018deep} extended the method in \cite{simonovsky2017dynamic} to segmentation tasks and proposed an efficient version, but their efficient version can only approximate depth-wise convolution instead of real convolution. Dynamic graph CNN \cite{wang2018dynamic} proposed a method that can dynamically updating the graph. \cite{xu2018spidercnn} presents a special family of filters to approximate the weight function instead of using MLPs. \cite{hermosilla2018monte} proposed a Monta Carlo approximation of 3D convolution by taking density into account. Our work differ from those in 3 aspects. Most importantly, our efficient version of a real convolution was never proposed in prior work. Also, we utilize density differently than \cite{hermosilla2018monte}, and we propose a deconvolution operator based on PointConv to perform semantic segmentation.

\section{PointConv}
We propose a convolution operation which extends traditional image convolution into the point cloud called \textbf{PointConv}. PointConv is an extension to the Monte Carlo approximation of the 3D continuous convolution operator. For each convolutional filter, it uses MLP to approximate a weight function, then applies a density scale to re-weight the learned weight functions. Sec.~\ref{DDC} introduces the structure of the PointConv layer. Sec.~\ref{FPD} introduces PointDeconv, using PointConv layers to deconvolve features.

\subsection{Convolution on 3D Point Clouds} \label{DDC}

Formally, convolution is defined as in Eq.(\ref{eq:conv1}) for functions $ f(\mathbf{x}) $ and $ g(\mathbf{x}) $ of a $d$-dimensional vector $ \mathbf{x} $.

\vspace{-0.2in}
\begin{align}
(f*g)(\mathbf{x}) = \iint\limits_{\boldsymbol{\tau} \in \mathbb{R}^d} f(\boldsymbol{\tau}) g(\mathbf{x} + \boldsymbol{\tau}) d\boldsymbol{\tau}  \label{eq:conv1}
\end{align}
\vskip -0.05in

\begin{figure}
	\centering
	\includegraphics[width=.25\textwidth]{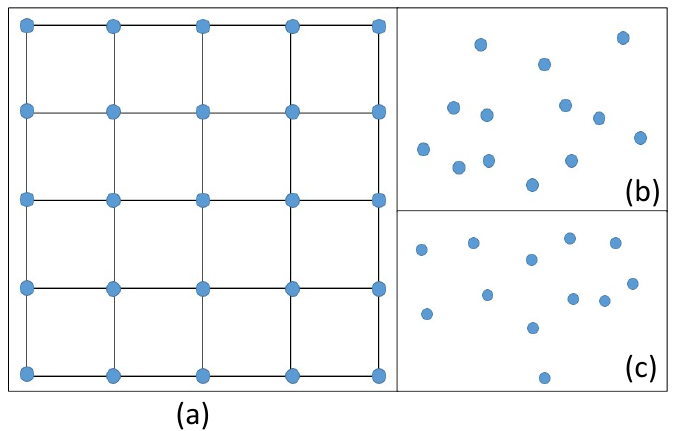}
	\caption{\textbf{Image grid vs. point cloud.} (a) shows a $ 5 \times 5 $ local region in a image, where the distance between points can only attain very few discrete values; (b) and (c) show that in different local regions within a point cloud, the order and the relative positions can be very different.}
	\label{fig1}
\vspace{-0.1in}
\end{figure}

Images can be interpreted as 2D discrete functions, which are usually represented as grid-shaped matrices. In CNN, each filter is restricted to a small local region, such as $3 \times 3, 5\times5$, etc. Within each local region, the relative positions between different pixels are always fixed, as shown in Figure \ref{fig1}(a). And the filter can be easily discretized to a summation with a real-valued weight for each location within the local region. 

A point cloud is represented as a set of 3D points $\{p_i | i = 1, ..., n\}$, where each point contains a position vector $(x, y, z)$ and its features such as color,  surface normal, etc. Different from images, point clouds have more flexible shapes. The coordinates $ p=(x, y, z) \in \mathbb{R}^3$ of a point in a point cloud are not located on a fixed grid but can take an arbitrary continuous value. Thus, the relative positions of different points are diverse in each local region. Conventional discretized convolution filters on raster images cannot be applied directly on the point cloud. Fig.~\ref{fig1} shows the difference between a local region in a image and a point cloud.

To make convolution compatible with point sets, we propose a permutation-invariant convolution operation, called \textbf{PointConv}. Our idea is to first go back to the continuous version of 3D convolution as:

\vspace{-0.2in}
\begin{small}
    \begin{align} \label{eq6}
        & Conv(W,F)_{xyz} = \nonumber \\
        &   \iiint\limits_{(\delta_x, \delta_y, \delta_z) \in G } W(\delta_x, \delta_y, \delta_z) F(x+\delta_x, y+\delta_y, z+\delta_z)d\delta_x \delta_y \delta_z
    \end{align}
\end{small}
\vskip -0.2in

\noindent where $F(x+\delta_x, y+\delta_y, z+\delta_z)$ is the feature of a point in the local region $G$ centered around point $p=(x, y, z)$. A point cloud can be viewed as a non-uniform sample from the continuous $\mathbb{R}^3 $ space. In each local region, $ (\delta_x, \delta_y, \delta_z) $ could be any possible position in the local region. We define \textit{PointConv} as the following:
 
\vspace{-0.2in}
\begin{small}
    \begin{align} \label{eq:pc_conv}
        & PointConv(S,W,F)_{xyz} = \nonumber \\
        &\sum_{(\delta_x, \delta_y, \delta_z) \in G } S(\delta_x, \delta_y, \delta_z) W(\delta_x, \delta_y, \delta_z) F(x+\delta_x, y+\delta_y, z+\delta_z)
    \end{align}
\end{small}
\vskip -0.15in

\begin{figure}
	\centering
	\includegraphics[width=.45\textwidth]{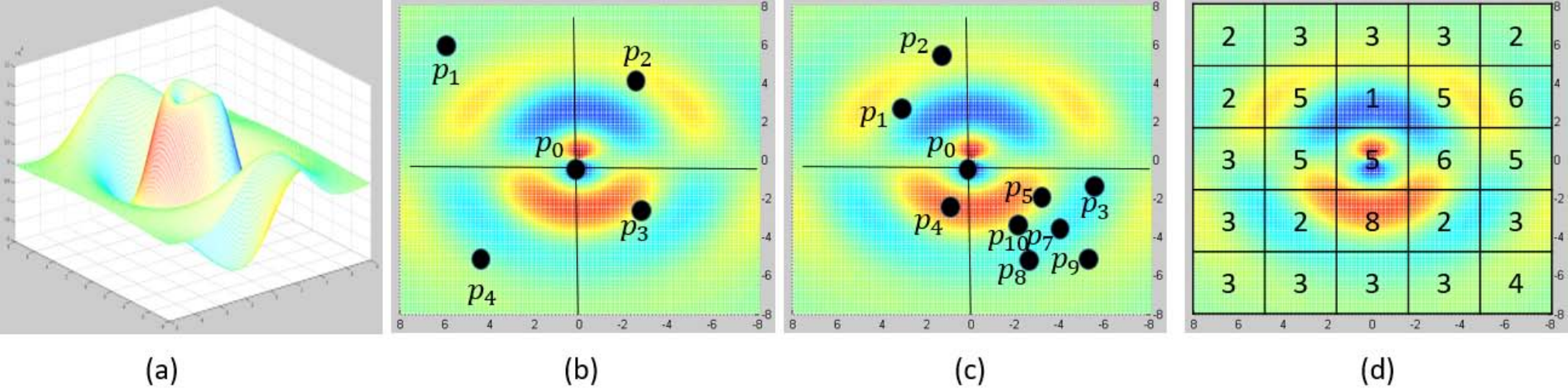}
	\caption{\textbf{2D weight function for PointConv.} (a) is a learned continuous weight function; (b) and (c) are different local regions in a 2d point cloud. Given 2d points, we can obtain the weights at particular locations. The same applies to 3D points. The regular discrete 2D convolution can be viewed as a discretization of the continuous convolution weight function, as in (d).}
	\label{fig3}
\vspace{-0.15in}
\end{figure}

\noindent where $S(\delta_x, \delta_y, \delta_z) $ is the inverse density at point $(\delta_x, \delta_y, \delta_z)$. $ S(\delta_x, \delta_y, \delta_z) $ is required because the point cloud can be sampled very non-uniformly\footnote{To see this, note the Monte Carlo estimate with a biased sample: $\int f(x) dx = \int \frac{f(x)}{p(x)} p(x) dx \approx \sum_i \frac{f(x_i)}{p(x_i)}$, for $x_i \sim p(x)$. Point clouds are often biased samples because many sensors have difficulties measuring points near plane boundaries, hence needing this reweighting}. Intuitively, the number of points in the local region varies across the whole point cloud, as in Figure \ref{fig3} (b) and (c). Besides, in Figure \ref{fig3} (c), points $ p_3, p_5, p_6, p_7, p_8, p_9, p_{10} $ are very close to one another, hence the contribution of each should be smaller. 

Our main idea is to approximate the weight function $ W(\delta_x, \delta_y, \delta_z) $ by multi-layer perceptrons from the 3D coordinates $(\delta_x, \delta_y, \delta_z)$ and and the inverse density $ S(\delta_x, \delta_y, \delta_z) $ by a kernelized density estimation \cite{turlach1993bandwidth} followed by a nonlinear transform implemented with MLP. Because the weight function highly depends on the distribution of input point cloud, we call the entire convolution operation \textit{PointConv}. \cite{jia2016dynamic,simonovsky2017dynamic} considered the approximation of the weight function but did not consider the approximation of the density scale, hence is not a full approximation of the continuous convolution operator. Our nonlinear transform on the density is also different from \cite{hermosilla2018monte}.

The weights of the MLP in PointConv are shared across all the points in order to maintain the permutation invariance. To compute the inverse density scale $S(\delta_x, \delta_y, \delta_z)$, we first estimate the density of each point in a point cloud offline using kernel density estimation (KDE), then feed the density into a MLP for a 1D nonlinear transform. The reason to use a nonlinear transform is for the network to decide adaptively whether to use the density estimates. 


\begin{figure*}
\begin{center}
\includegraphics[width=0.85\textwidth]{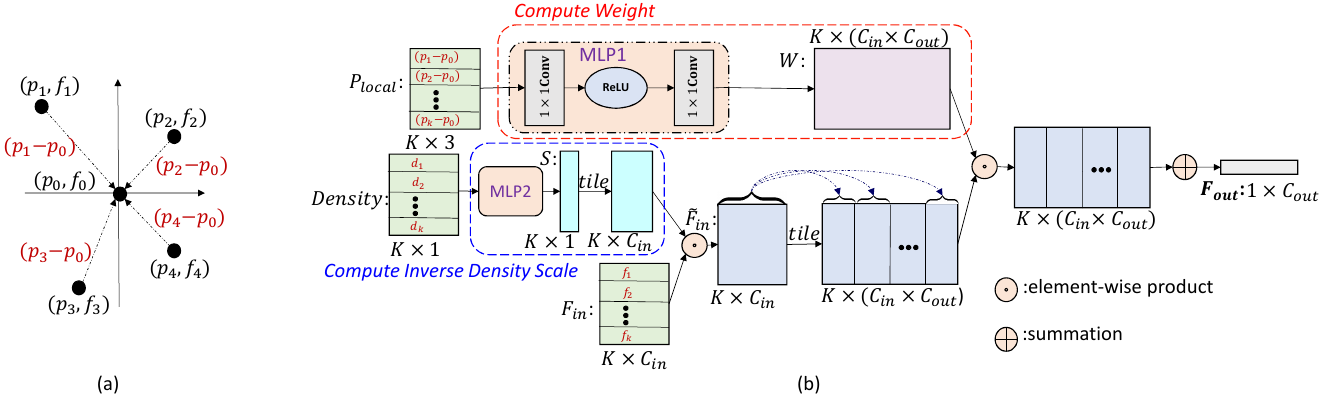}
\end{center}
   \caption{\textbf{PointConv}. (a) shows a local region with the coordinates of points transformed from global into local coordinates, $p$ is the coordinates of points, and $f$ is the corresponding feature; (b) shows the process of conducting PointConv on one local region centered around one point $(p_0,f_0)$. The input features come form the K nearest neighbors centered at $(p_0, f_0)$, and the output feature is $\mathbf{F}_{out}$ at $p_0$.}
\label{fig4}
\vskip -0.15in
\end{figure*}

Figure \ref{fig4} shows the PointConv operation on a $K$-point local region. Let $C_{in}, C_{out}$ be the number of channels for the input feature and output feature, $k, c_{in}, c_{out}$ are the indices for $k$-th neighbor, $c_{in}$-th channel for input feature and $c_{out}$-th channel for output feature. The inputs are the 3D local positions of the points $ P_{local} \in \mathbb{R}^{K\times 3}$, which can be computed by subtracting the coordinate of the centroid of the local region and the feature $ F_{in} \in \mathbb{R}^{K\times C_{in}} $ of the local region. We use $1\times 1$ convolution to implement the MLP. The output of the weight function is $W \in \mathbb{R}^{K \times (C_{in} \times C_{out})}$. So, $\mathbf{W}(k, c_{in}) \in \mathbb{R}^{C_{out}}$ is a vector. The density scale is $S \in \mathbb{R}^{K}$. After convolution, the feature $F_{in}$ from a local region with $K$ neighbour points are encoded into the output feature $\mathbf{F}_{out} \in \mathbb{R}^{C_{out}}$, as shown in Eq.(\ref{eq:PointConv}).
\vspace{-0.05in}
\begin{equation} \label{eq:PointConv}
\mathbf{F}_{out} = \sum^{K}_{k=1}\sum^{C_{in}}_{c_{in} = 1} S(k)\mathbf{W}(k, c_{in})F_{in}(k, c_{in})
\end{equation}
\vskip -0.05in

PointConv learns a network to approximate the continuous weights for convolution. For each input point, we can compute the weights from the MLPs using its relative coordinates. 
Figure \ref{fig3}(a) shows an example continuous weight function for convolution. 
With a point cloud input as a discretization of the continuous input, a discrete convolution can be computed by Fig.~\ref{fig3}(b) to extract the local features, which would work (with potentially different approximation accuracy) for different point cloud samples (Figure \ref{fig3}(b-d)), including a regular grid (Figure \ref{fig3}(d)). Note that in a raster image, the relative positions in local region are fixed. Then PointConv (which takes only relative positions as input for the weight functions) would output the same weight and density across the whole image, where it reduces to the conventional discretized convolution. 

In order to aggregate the features in the entire point set, we use a hierarchical structure that is able to combine detailed small region features into abstract features that cover a larger spatial extent. The hierarchical structure we use is composed by several \textit{feature encoding modules}, which is similar to the one used in PointNet++ \cite{qi2017pointnet++}. Each module is roughly equivalent to one layer in a convolutional CNN. The key layers in each \textit{feature encoding module} are the sampling layer, the grouping layer and the PointConv.

The drawback of this approach is that each filter needs to be approximated by a network, hence is very inefficient. In Sec.\ref{IMP}, we propose an efficient approach to implement PointConv.

\subsection{Feature Propagation Using Deconvolution} \label{FPD}

For the segmentation task, we need point-wise prediction. In order to obtain features for all the input points, an approach to propagate features from a subsampled point cloud to a denser one is needed. PointNet++ \cite{qi2017pointnet++} proposes to use distance-based interpolation to propagate features, which is reasonable due to local correlations inside a local region. However, this does not take full advantage of the deconvolution operation that captures local correlations of propagated information from the coarse level. We propose to add a PointDeconv layer based on the PointConv, as a deconvolution operation to address this issue. 
\begin{figure}
	\centering
	\includegraphics[width=.5\textwidth]{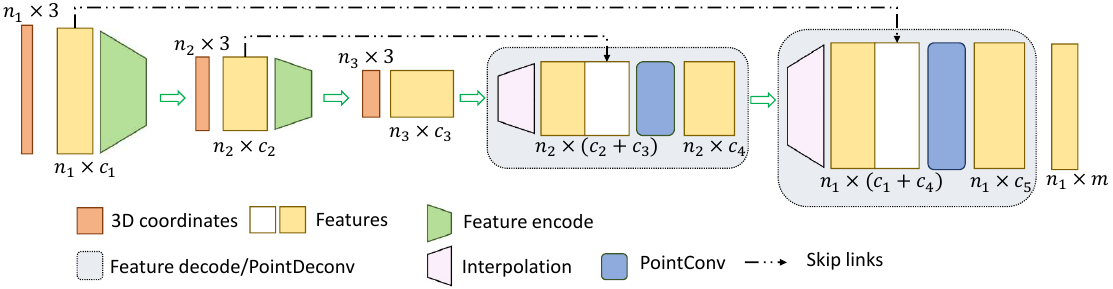}
	\caption{\textbf{Feature encoding and propagation.} This figure shows how the features are encoded and propagated in the network for a $m$ classes segmentation task. $n$ is the number of points in each layer, $c$ is the channel size for the features. Best viewed in color.}
	\label{fig5}
	\vskip -0.1in
\end{figure}

As shown in Fig.~\ref{fig5}, PointDeconv is composed of two parts: interpolation and PointConv. Firstly, we employ an interpolation to propagate coarse features from previous layer. Following \cite{qi2017pointnet++}, the interpolation is conducted by linearly interpolating features from the 3 nearest points. Then, the interpolated features are concatenated with features from the convolutional layers with the same resolution using skip links. After concatenation, we apply PointConv on the concatenated features to obtain the final deconvolution output,  similar to the image deconvolution layer \cite{noh2015learning}. We apply this process until the features of all the input points have been propagated back to the original resolution.

\section{Efficient PointConv} \label{IMP}

The naive implementation of the PointConv is memory consuming and inefficient. Different from \cite{simonovsky2017dynamic}, we propose a novel reformulation to implement PointConv by reducing it to two standard operations: matrix multiplication and 2d convolution. This novel trick not only takes advantage of the parallel computing of GPU, but also can be easily implemented using main-stream deep learning frameworks. Because the inverse density scale does not have such memory issues, the following discussion mainly focuses on the weight function.

Specifically, let $B$ be the mini-batch size in the training stage, $N$ be the number of points in a point cloud, $K$ be the number of points in each local region, $C_{in}$ be the number of input channels, and $C_{out}$ be the number of output channels. For a point cloud, each local region shares the same weight functions which can be learned using MLP. However, weights computed from the weight functions at different points are different. The size of the weights filters generated by the MLP is $B \times N \times K \times (C_{in} \times C_{out})$. Suppose $B = 32$, $N = 512$, $ K = 32$, $C_{in} = 64$, $C_{out} = 64$, and the filters are stored with single point precision. Then, the memory size for the filters is $8 GB$ for only one layer. The network would be hard to train with such high memory consumption. \cite{simonovsky2017dynamic} used very small network with few filters which significantly degraded its performance. To resolve this problem, we propose a memory efficient version of PointConv based on the following lemma:

\vspace{-0.1in}
\begin{lemma}
\label{le:me}
The PointConv is equivalent to the following formula:
$\mathbf{F}_{out} = Conv_{1\times 1}(\mathbf{H}, (\mathbf{S} \cdot \mathbf{{F}}_{in})^T \otimes \mathbf{M})$
where $\mathbf{M} \in \mathbb{R}^{K\times C_{mid}}$ is the input to the last layer in the MLP for computing the weight function, and $\mathbf{H} \in \mathbb{R}^{C_{mid} \times (C_{in} \times C_{out})}$ is the weights of the last layer in the same MLP, $Conv_{1\times 1}$ is $1\times 1$ convolution.
\vspace{-0.1in}
\end{lemma}

\noindent \textbf{Proof:} Generally, the last layer of the MLP is a linear layer. In one local region, let $\mathbf{\widetilde{F}_{in}} = \mathbf{S}\cdot \mathbf{F_{in}} \in \mathbb{R}^{K\times C_{in}}$ and rewrite the MLP as a $1\times1$ convolution so that the output of the weight function is $\mathbf{W} = Conv_{1\times1}(H, M) \in \mathbb{R}^{K\times (C_{in} \times C_{out})}$. Let $k$ is the index of the points in a local region, and $c_{in}, c_{mid}, c_{out}$ are the indices of the input, middle layer and the filter output, respectively. Then $\mathbf{W}(k, c_{in}) \in \mathbb{R}^{C_{out}}$ is a vector from $\mathbf{W}$. And the $\mathbf{H}(c_{mid}, c_{in}) \in \mathbb{R}^{C_{out}}$ is a vector from $\mathbf{H}$. According to Eq.(\ref{eq:PointConv}), the PointConv can be expressed in Eq.(\ref{eq:pointconv_ori}).

\vspace{-0.2in}
\begin{align}
\mathbf{F}_{out} = \sum^{K-1}_{k=0}\sum^{C_{in}-1}_{c_{in}=0}(\mathbf{W}(k, c_{in})\mathbf{\widetilde{F}_{in}}(k, c_{in})) \label{eq:pointconv_ori} 
\end{align}

Let's explore Eq.(\ref{eq:pointconv_ori}) in a more detailed manner. The output of the weight function can be expressed as:

\vspace{-0.2in}
\begin{align}
\mathbf{W}(k, c_{in}) &= \sum^{C_{mid} - 1}_{c_{mid} = 0}(\mathbf{M}(k, c_{mid})\mathbf{H}(c_{mid}, c_{in})) \label{eq:weightnet}
\end{align}

Substituting Eq.(\ref{eq:weightnet}) into Eq.(\ref{eq:pointconv_ori}).

\begin{figure*}
    \centering
    \includegraphics[width=.7\textwidth, height=.27\textwidth]{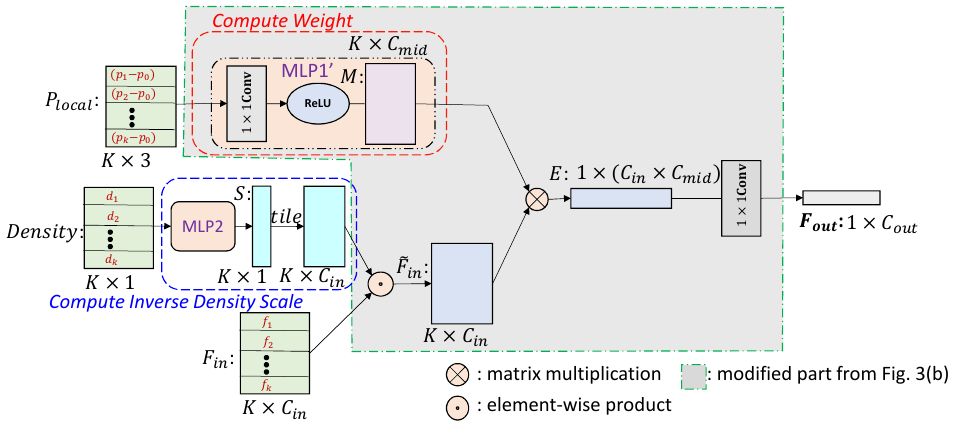}
    \caption{\textbf{Efficient PointConv.} The memory efficient version of PointConv on one local region with $K$ points.}
    \label{fig:PointConv_v2}
    \vskip -0.15in
\end{figure*}

\vspace{-0.1in}
\begin{scriptsize}
\begin{align}
\mathbf{F}_{out} & = \sum^{K-1}_{k=0}\sum^{C_{in} -1}_{c_{in}=0}(\mathbf{\widetilde{F}_{in}}(k, c_{in})\sum^{C_{mid} - 1}_{c_{mid} = 0}(\mathbf{M}(k, c_{mid})\mathbf{H}(c_{mid}, c_{in}))) \nonumber\\
& = \sum^{C_{in} -1}_{c_{in}=0}\sum^{C_{mid} - 1}_{c_{mid} = 0}(\mathbf{H}(c_{mid}, c_{in})\sum^{K-1}_{k=0}(\mathbf{\widetilde{F}_{in}}(k, c_{in})\mathbf{M}(k, c_{mid}))) \nonumber\\
& = Conv_{1\times 1}(\mathbf{H}, \mathbf{\widetilde{F}_{in}}^T \mathbf{M}) \label{eq:pointconv_v2_4}
\end{align}
\end{scriptsize}
\vskip -0.2in
Thus, the original PointConv can be equivalently reduced to a matrix multiplication and a $1\times 1$ convolution. Figure \ref{fig:PointConv_v2} shows the efficient version of PointConv.

In this method, instead of storing the generated filters in memory, we divide the weights filters into two parts: the intermediate result $M$ and the convolution kernel $H$. As we can see, the memory consumption reduces to $\mathbf{\frac{C_{mid}}{K \times C_{out}}}$ of the original version. With the same input setup as the Figure \ref{fig4} and let $C_{mid} = 32$, the memory consumption is $0.1255 GB$, which is about $1/64$ of the original PointConv. 
\section{Experiments}

In order to evaluate our new PointConv network, we conduct experiments on several widely used datasets, ModelNet40 \cite{wu20153d},  ShapeNet  \cite{chang2015shapenet} and  ScanNet \cite{dai2017scannet}. In order to demonstrate that our PointConv is able to fully approximate conventional convolution, we also report results on the CIFAR-10 dataset \cite{krizhevsky2009learning}. In all experiments, we implement the models with Tensorflow on a GTX 1080Ti GPU using the Adam optimizer. ReLU and batch normalization are applied after each layer except the last fully connected layer. 

\subsection{Classification on ModelNet40}

ModelNet40 contains 12,311 CAD models from 40 man-made object categories. We use the official split with 9,843 shapes for training and 2,468 for testing. Following the configuration in \cite{qi2017pointnet}, we use the source code for PointNet~\cite{qi2017pointnet} to sample 1,024 points uniformly and compute the normal vectors from the mesh models. For fair comparison, we employ the same data augmentation strategy as \cite{qi2017pointnet} by randomly rotating the point cloud along the $z$-axis and jittering each point by a Gaussian noise with zero mean and 0.02 standard deviation. In Table \ref{table:modelnet40},  PointConv achieved state-of-the-art performance among methods based on 3D input. ECC\cite{simonovsky2017dynamic} which is similar to our approach, cannot scale to a large network, which limited their performance.

\setlength{\tabcolsep}{4pt}
\begin{table}
	\begin{center}
		\caption{\textbf{ModelNet40 Classification Accuracy}}
		\label{table:modelnet40}
		\begin{tabular}{c|l|c}
			\hline\noalign{\smallskip}
			 Method & Input & Accuracy(\%) \\
			\noalign{\smallskip}
			\hline
			\noalign{\smallskip}
			Subvolume \cite{qi2016volumetric} & voxels & 89.2 \\
			ECC \cite{simonovsky2017dynamic} & graphs & 87.4 \\
			Kd-Network \cite{klokov2017escape} & 1024 points & 91.8 \\
			PointNet \cite{qi2017pointnet} & 1024 points & 89.2 \\
			PointNet++ \cite{qi2017pointnet++} & 1024 points & 90.2 \\
			PointNet++ \cite{qi2017pointnet++} & 5000 points+normal & 91.9 \\
			SpiderCNN \cite{xu2018spidercnn} & 1024 points+normal & 92.4 \\
			\noalign{\smallskip}
			\hline
			\noalign{\smallskip}
            PointConv & 1024 points+normal & \textbf{92.5} \\
			\hline
		\end{tabular}
	\end{center}
	\vskip -0.1in
\end{table}
\setlength{\tabcolsep}{1.4pt}

\subsection{ShapeNet Part Segmentation}

\begin{figure}
	\centering
	\includegraphics[width=.3\textwidth]{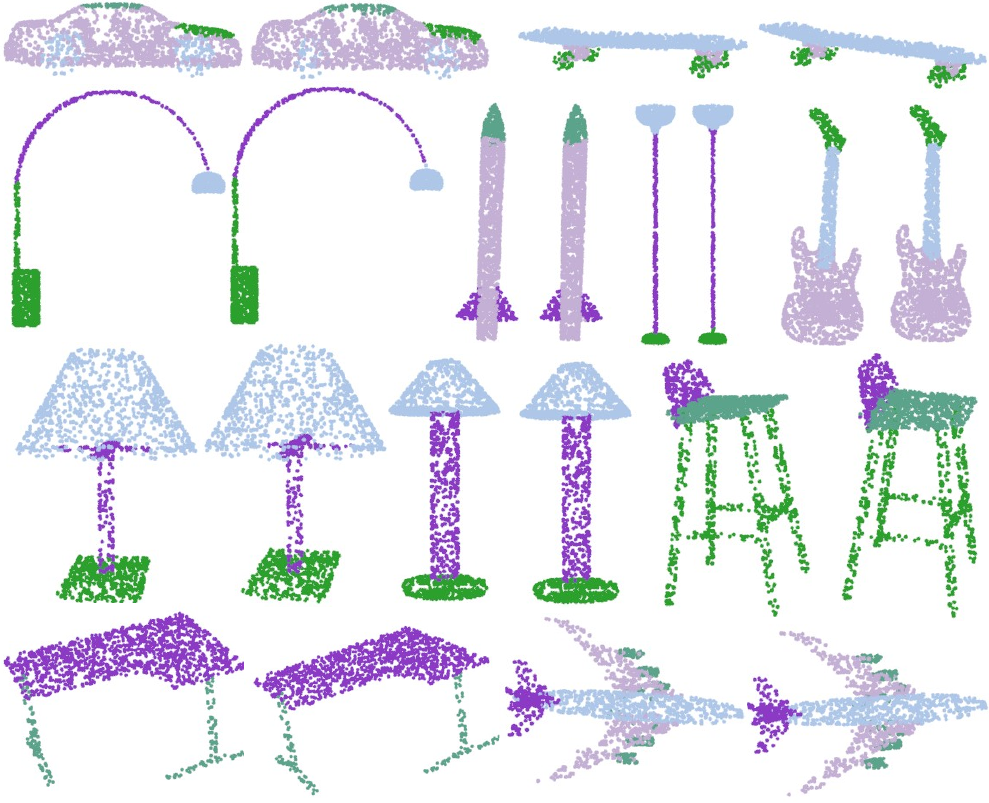}
	\caption{\textbf{Part segmentation results.} For each pair of objects, the left one is the ground truth, the right one is predicted by PointConv. Best viewed in color.}
	\label{fig6}
\vspace{-0.3in}
\end{figure}

Part segmentation is a challenging fine-grained 3D recognition task. The ShapeNet dataset contains 16,881 shapes from 16 classes and 50 parts in total. The input of the task is shapes represented by a point cloud, and the goal is to assign a part category label to each point in the point cloud. The category label for each shape is given. We follow the experiment setup in most related work \cite{qi2017pointnet++, su2018splatnet, xu2018spidercnn, klokov2017escape}. It is common to narrow the possible part labels to the ones specific to the given object category by using the known input 3D object category. And we also compute the normal direction on each point as input features to better describe the underlying shape. Figure \ref{fig6} visualizes some sample results.

\setlength{\tabcolsep}{4pt}
\begin{table}
\vspace{0.2in}
	\begin{center}
		\caption{\textbf{Results on ShapeNet part dataset.} Class avg. is the mean IoU averaged across all object categories, and inctance avg. is the mean IoU across all objects.}
		\label{table1}
		\begin{tabular}{l|cc}
			\hline\noalign{\smallskip}
			      &class avg. & instance avg.\\
			\noalign{\smallskip}
			\hline
			\noalign{\smallskip}
			SSCNN \cite{yi2017syncspeccnn} & 82.0 & 84.7\\
			Kd-net \cite{klokov2017escape} & 77.4 & 82.3\\
			PointNet \cite{qi2017pointnet}       & 80.4 & 83.7\\
			PointNet++\cite{qi2017pointnet++}  & 81.9 & 85.1\\
			SpiderCNN \cite{xu2018spidercnn}& 82.4 & 85.3\\
			SPLATNet$_{3D}$ \cite{su2018splatnet}& 82.0 & 84.6\\
			SSCN \cite{graham2017submanifold}& - & \textbf{86.0}\\
			\noalign{\smallskip}
			\hline
			\noalign{\smallskip}
			PointConv & \textbf{82.8} & 85.7\\
			\noalign{\smallskip}
			\hline
		\end{tabular}
	\end{center}
\vspace{-0.4in}
\end{table}
\setlength{\tabcolsep}{1.4pt}

We use point intersection-over-union(IoU) to evaluate our PointConv network, same as PointNet++ \cite{qi2017pointnet++}, SPLATNet \cite{su2018splatnet} and some other part segmentation algorithms \cite{yi2017syncspeccnn, klokov2017escape,xu2018spidercnn, graham2017submanifold}. The results are shown in Table \ref{table1}. PointConv obtains a class average mIoU of 82.8\% and an instance average mIoU of 85.7\%, which are on par with the state-of-the-art algorithms which only take point clouds as input. According to \cite{su2018splatnet}, the SPLATNet$_{2D-3D}$ also takes rendered 2D views as input. Since our PointConv only takes 3D point clouds as input, for fair comparison, we only compare our result with the SPLATNet$_{3D}$ in \cite{su2018splatnet}.

\subsection{Semantic Scene Labeling}

Datasets such as ModelNet40 \cite{wu20153d} and ShapeNet \cite{chang2015shapenet} are man-made synthetic datasets. As we can see in the previous section, most state-of-the-art algorithms are able to obtain relatively good results on such datasets. To evaluate the capability of our approach in processing realistic point clouds, which contains a lot of noisy data, we evaluate our PointConv on semantic scene segmentation using the ScanNet dataset. The task is to predict semantic object labels on each 3D point given indoor scenes represented by point clouds. The newest version of ScanNet \cite{dai2017scannet} includes updated annotations for all 1513 ScanNet scans and 100 new test scans with all semantic labels publicly unavailable and we submitted our results to the official evaluation server to compare against other approaches. 

\begin{figure}
	\centering
	\includegraphics[width=.4\textwidth]{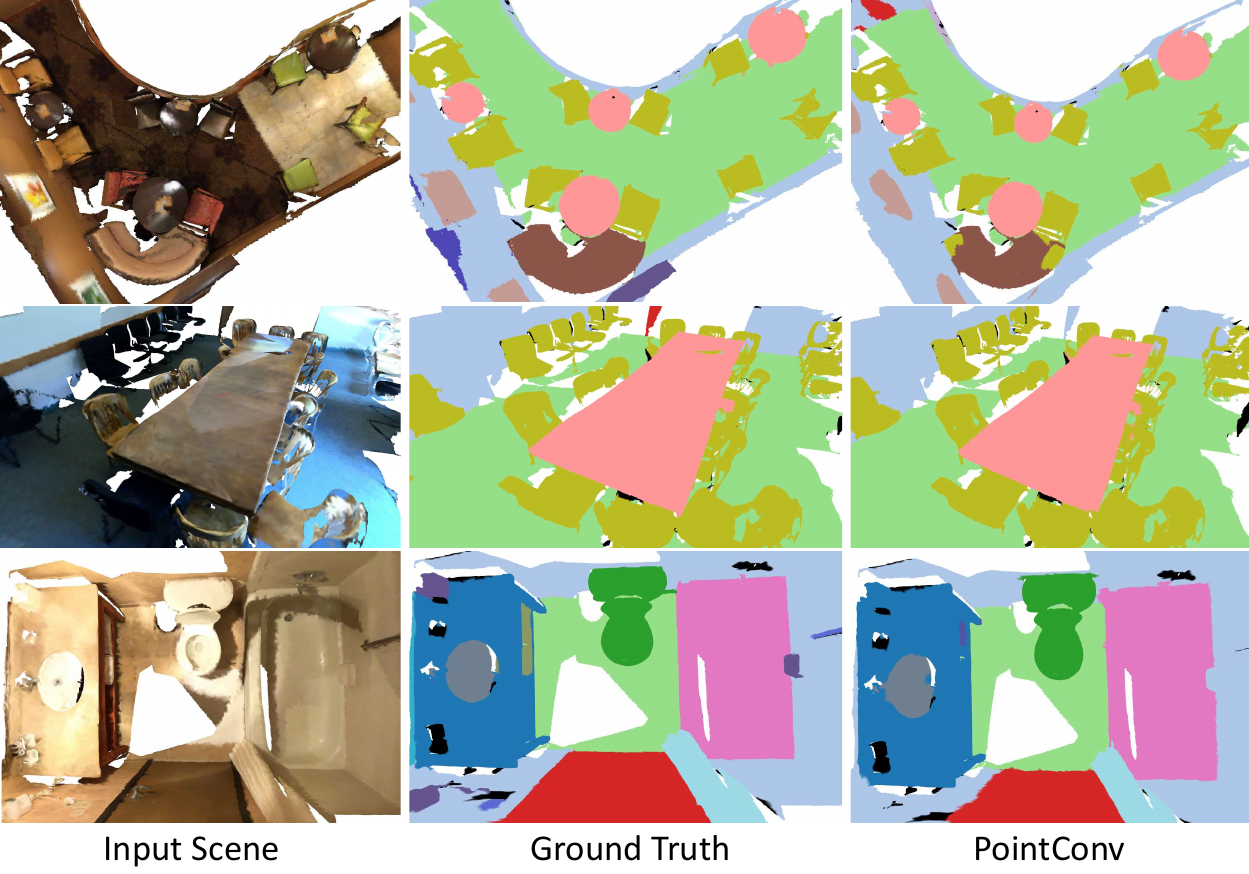}
	\caption{\textbf{Examples of semantic scene labeling.} The images from left to right are the input scenes, the ground truth segmentation, and the prediction from PointConv. For better visualization, the point clouds are converted into mesh format. Best viewed in color.}
	\label{fig7}
\vspace{-0.05in}
\end{figure}

We compare our algorithm with Tangent Convolutions \cite{tatarchenko2018tangent}, SPLAT Net \cite{su2018splatnet}, PointNet++ \cite{qi2017pointnet++} and ScanNet \cite{dai2017scannet}. All the algorithm mentioned reported their results on the new ScanNet dataset to the benchmark, and the inputs of the algorithms only uses 3D coordinates data plus RGB. In our experiments, we generate training samples by randomly sample $3m \times 1.5m \times 1.5m$ cubes from the indoor rooms, and evaluate using a sliding window over the entire scan. We report intersection over union (IoU) as our main measures, which is the same as the benchmark. We visualize some example semantic segmentation results in Figure \ref{fig7}. The mIoU is reported in Table \ref{table2}. The mIoU is the mean of IoU across all the categories. Our PointConv outperforms other algorithm by a significant margin (Table \ref{table2}). The total running time of PointConv for training one epoch on ScanNet on one GTX1080Ti is around 170s, and the evaluation time with 8 $\times$ 8192 points is around 0.5s.

\setlength{\tabcolsep}{4pt}
\begin{table}
	\begin{center}
		\caption{\textbf{Semantic Scene Segmentation results on ScanNet}}
		\label{table2}
		\begin{tabular}{ll}
			\hline\noalign{\smallskip}
			Method & mIoU(\%)\\
			\noalign{\smallskip}
			\hline
			\noalign{\smallskip}
			ScanNet \cite{dai2017scannet} & 30.6 \\
			PointNet++ \cite{qi2017pointnet++} & 33.9\\
			SPLAT Net \cite{su2018splatnet} & 39.3\\
			Tangent Convolutions \cite{tatarchenko2018tangent} & 43.8\\
			\noalign{\smallskip}
			\hline
			\noalign{\smallskip}
			PointConv & \textbf{55.6}\\
			\hline
		\end{tabular}
	\end{center}
\vspace{-0.2in}
\end{table}
\setlength{\tabcolsep}{1.4pt}

\subsection{Classification on CIFAR-10}

In Sec.\ref{DDC}, we claimed that PointConv can be equivalent with 2D CNN. If this is true, then the performance of a network based on PointConv should be equivalent to that of a raster image CNN. In order to verify that, we use the CIFAR-10 dataset as a comparison benchmark. We treat each pixel in CIFAR-10 as a 2D point with $xy$ coordinates and RGB features. The point clouds are scaled onto the unit ball before training and testing.

Experiments show that PointConv on CIFAR-10 indeed has the same learning capacities as a 2D CNN. Table \ref{table:alexnet} shows the results of image convolution and PointConv. From the table, we can see that the accuracy of PointCNN\cite{li2018pointcnn} on CIFAR-10 is only $80.22 \%$, which is much worse than image CNN. However, for 5-layer networks, the network using PointConv is able to achieve $89.13 \%$, which is similar to the network using image convolution. And, PointConv with VGG19 \cite{simonyan2014very} structure can also achieve on par accuracy comparing with VGG19.

\setlength{\tabcolsep}{4pt}
\begin{table}
	\begin{center}
		\caption{\textbf{CIFAR-10 Classification Accuracy}}
		\label{table:alexnet}
		\begin{tabular}{c|c}
			\hline\noalign{\smallskip}
			    & Accuracy(\%)\\
			\noalign{\smallskip}
			\hline
			\noalign{\smallskip}
			Image Convolution & 88.52 \\
            AlexNet \cite{krizhevsky2012imagenet} & 89.00\\
            VGG19 \cite{simonyan2014very} & 93.60 \\
            PointCNN \cite{li2018pointcnn} & 80.22 \\
            SpiderCNN \cite{xu2018spidercnn} & 77.97 \\
            \noalign{\smallskip}
			\hline
			\noalign{\smallskip}
            PointConv(5-layer) & 89.13 \\
            PointConv(VGG19) & 93.19 \\
			\hline
		\end{tabular}
	\end{center}
\vspace{-0.3in}
\end{table}
\setlength{\tabcolsep}{1.4pt}

\section{Ablation Experiments and Visualizations}

In this section, we conduct additional experiments to evaluate the effectiveness of each aspect of PointConv. Besides the ablation study on the structure of the PointConv, we also give an in-depth breakdown on the performance of PointConv on the ScanNet dataset. Finally, we  provide some learned filters for visualization.

\subsection{The Structure of MLP}

In this section, we design experiments to evaluate the choice of MLP parameters in PointConv. For fast evaluation, we generate a subset from the ScanNet dataset as a classification task. Each example in the subset is randomly sampled from the original scene scans with 1,024 points. There are 20 different scene types for the ScanNet dataset. The reason why we use a subset of ScanNet dataset is we want to avoid fitting the wrong parameters on a dataset that is too simple such as ModelNet40. The selected dataset is a realistic 3D point cloud with RGB information. Since the problem is complex enough, we could imagine parameters that are good enough for other datasets with similar complexity.

\begin{figure*}
	\centering
	\includegraphics[width=0.8\textwidth]{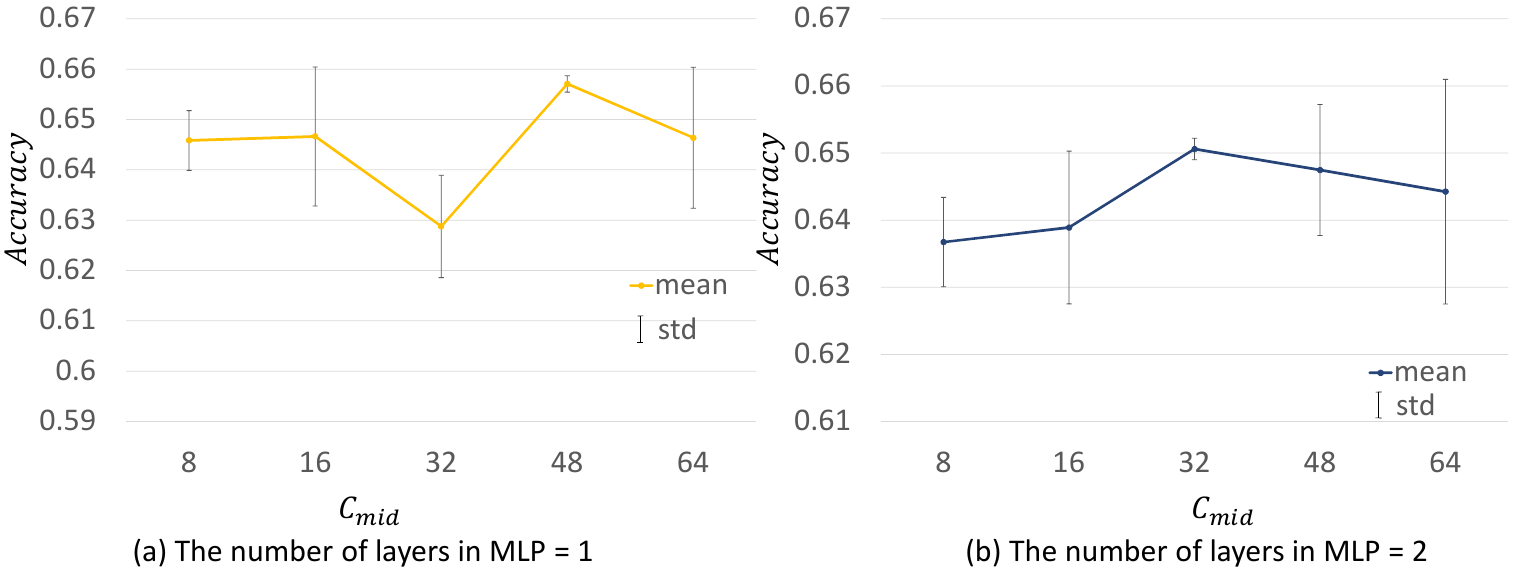}
	\caption{Classification accuracy of different choice of $C_{mid}$ and layers number of MLP.}
	\label{fig:cmid_mlp}
\end{figure*}

We empirically sweep over different choices of $C_{mid}$ and different number of layers of the MLP in PointConv. Each experiment was conducted for 3 random trials. The results is shown in Figure \ref{fig:cmid_mlp}. From the results, we find that larger $C_{mid}$ does not necessarily give better classification results. And the different number of layers in MLP does not give much difference in classification results. Since $C_{mid}$ is linearly correlated with the memory consumption of each PointConv layer, this results shows that we can choose a reasonably small $C_{mid}$ for greater memory efficiency.

\subsection{Inverse Density Scale}

In this section, we study the effectiveness of the inverse density scale $S$. We choose ScanNet as our evaluation task since the point clouds in ScanNet are generated from real indoor scenes. We follow the standard training/validation split provided by the authors. We train the network with and without the inverse density scale as described in Sec.~\ref{DDC}, respectively. Table \ref{table:eval_stride} shows the results. As we can see, PointConv with inverse density scale performs better than the one without by about $1\%$, which proves the effectiveness of inverse density scale. In our experiments, we observe that inverse density scale tend to be more effective in layers closer to the input. In deep layers, the MLP tends to diminish the effect of the density scale. One possible reason is that with farthest point sampling algorithm as our sub-sampling algorithm, the point cloud in deeper layer tend to be more uniformly distributed. And as shown in Table \ref{table:eval_stride}, directly applying density without using the nonlinear transformation gives worse result comparing with the one without density on ScanNet dataset, which shows that the nonlinear transform is able to learn the inverse density scale in the dataset. 

\subsection{Ablation Studies on ScanNet}

As one can see, our PointConv outperforms other approaches with a large margin. Since we are only allowed to submit one final result of our algorithm to the benchmark server of ScanNet, we perform more ablation studies for PointConv using the public validation set provide by \cite{dai2017scannet}. For the segmentation task, we train our PointConv with 8,192 points randomly sampled from a $3m \times 1.5m \times 1.5m$, and evaluate the model with exhaustively choose all points in the $3m \times 1.5m \times 1.5m$ cube in a sliding window fashion through the xy-plane with different stride sizes. For robustness, we use a majority vote from 5 windows in all of our experiments. From Table \ref{table:eval_stride}, we can see that smaller stride size is able to improve the segmentation results, and the RGB information on ScanNet does not seem to significantly improve the segmentation results. Even without these additional improvements, PointConv still outperforms baselines by a large margin.
\vspace{-0.05in}

\setlength{\tabcolsep}{4pt}
\begin{table}[htb]
	\begin{center}
		\caption{\textbf{Ablation study on ScanNet.} With and without RGB information, inverse density scale and using different stride size of sliding window.}
		\label{table:eval_stride}
		\begin{tabular*}{0.45\textwidth}{c|c|c|c|c}
			\hline
			  & Stride   &  & mIoU & mIoU\\
			Input & Size(m) & mIoU & No Density & Density \\
			  &   & &   & (no MLP) \\
			\hline
			\multirow{3}{*}{xyz} & 0.5 & 61.0 & 60.3 & 60.1\\
                                 & 1.0 & 59.0 & 58.2 & 57.7\\
                                 & 1.5 & 58.2 & 56.9 & 57.3\\
			\hline
            \multirow{3}{*}{xyz+RGB} & 0.5 & 60.8 & 58.9 & -\\
                                 & 1.0 & 58.6 & 56.7 & -\\
                                 & 1.5 & 57.5 & 56.1 & -\\
			\hline
		\end{tabular*}
	\end{center}
\vspace{-0.25in}
\end{table}
\setlength{\tabcolsep}{1.4pt}

\subsection{Visualization}

\begin{figure}
\centering
\includegraphics[width=.45\textwidth]{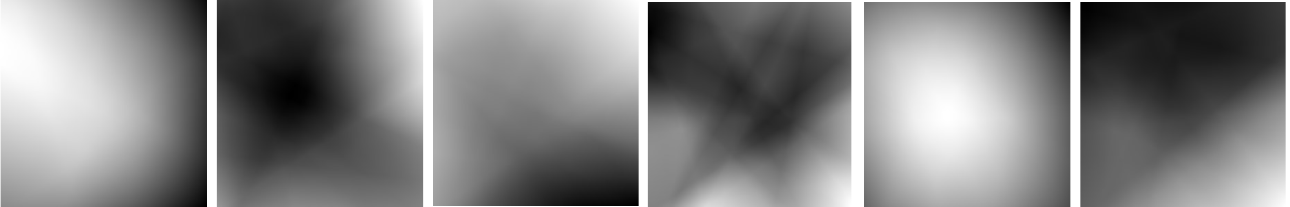}
\caption{\textbf{Learned Convolutional Filters.} The convolution filters learned by the MLPs on ShapeNet.For better visualization, we take all weights filters from $z=0$ plane.}
\label{fig:weights}
\vskip -0.15in
\end{figure}

Figure \ref{fig:weights} visualizes the learned filters from the MLPs in our PointConv. In order to better visualize the filters, we sample the learned functions through a plane $z = 0$. From the Figure \ref{fig:weights}, we can see some patterns in the learned continuous filters.

\section{Conclusion}

In this work, we proposed a novel approach to perform convolution operation on 3D point clouds, called PointConv. PointConv trains multi-layer perceptrons on local point coordinates to approximate continuous weight and density functions in convolutional filters, which makes it naturally permutation-invariant and translation-invariant. This allows deep convolutional networks to be built directly on 3D point clouds. We proposed an efficient implementation of it which greatly improved its scalability. We demonstrated its strong performance on multiple challenging benchmarks and capability of matching the performance of a grid-based convolutional network in 2D images. In future work, we would like to adopt more mainstream image convolution network architectures into point cloud data using PointConv, such as ResNet and DenseNet. The code can be found here: https://github.com/DylanWusee/pointconv.

\nocite{chang2015shapenet,dai2017scannet,jacobsen2016structured,noh2015learning,qi2016volumetric,qi2017frustum}
\nocite{qi2017pointnet,qi2017pointnet++,ravanbakhsh2016deep,su2015multi,yi2017syncspeccnn,zhou2017voxelnet}
\nocite{bronstein2010scale,chen2003visual,fang20153d,gressin2013towards,guo20153d,he2016deep,kingma2014adam}
\nocite{krizhevsky2012imagenet,ling2007shape,maturana2015voxnet,rusu2009fast,simonyan2014very,wu2014interactive}
\nocite{wu20153d, huang2018recurrent, qi20173d, zhou2017unsupervised, chu2018surfconv}

{\small
\bibliographystyle{ieee}
\bibliography{egbib}
}

\end{document}